\documentclass[11pt,a4paper]{article}
\usepackage[hyperref]{acl2019}
\usepackage{times}
\usepackage{latexsym}
\usepackage[labelfont=bf]{caption}

\usepackage{amssymb}
\usepackage{amsmath}
\usepackage{enumitem}
\usepackage{pifont}
\usepackage{makecell}
\usepackage{url}
\usepackage{graphicx}
\usepackage{footnote}
\usepackage{tablefootnote}
\usepackage{multirow}
\usepackage{booktabs}  
\usepackage{listings}
\usepackage{float}
\usepackage{stfloats}

\makesavenoteenv{tabular}
\makesavenoteenv{table}
\graphicspath{{images/}}

\aclfinalcopy 
\def\aclpaperid{66} 

\newcommand{\xmark}{\ding{55}}%
\DeclareMathOperator*{\argmax}{arg\,max}

\newcommand{\hide}[1]{}

\title{Open Sesame: Getting Inside BERT's Linguistic Knowledge}

\author{Yongjie Lin$\*^{a,}$\footnotemark[1] \and Yi Chern Tan$\*^{a,}$\footnotemark[1] \and Robert Frank$\*^{b}$ \\
  $\*^a$Department of Computer Science, Yale University \\
  $\*^b$Department of Linguistics, Yale University \\
  \texttt{\{yongjie.lin, yichern.tan, robert.frank\}@yale.edu} \\}

\date{}

\begin{document}
\maketitle

{\renewcommand{\thefootnote}{\fnsymbol{footnote}} \stepcounter{footnote}\footnotetext{Equal contribution.}}
\setcounter{footnote}{0}

\begin{abstract}

How and to what extent does BERT encode syntactically-sensitive hierarchical information or positionally-sensitive linear information? Recent work has shown that contextual representations like BERT perform well on tasks that require sensitivity to linguistic structure. We present here two studies which aim to provide a better understanding of the nature of BERT's representations. The first of these focuses on the identification of structurally-defined elements using diagnostic classifiers, while the second explores BERT's representation of subject-verb agreement and anaphor-antecedent dependencies through a quantitative assessment of self-attention vectors. In both cases, we find that BERT encodes positional information about word tokens well on its lower layers, but switches to a hierarchically-oriented encoding on higher layers. We conclude then that BERT's representations do indeed model linguistically relevant aspects of hierarchical structure, though they do not appear to show the sharp sensitivity to hierarchical structure that is found in human processing of reflexive anaphora.\footnote{The code is available at \url{https://github.com/yongjie-lin/bert-opensesame}.}

\end{abstract}

\section{Introduction}

Word embeddings have become an important cornerstone in any NLP pipeline. Although such embeddings traditionally involve context-free distributed representations of words \cite{mikolov2013distributed, pennington2014glove}, recent successes with contextualized representations \cite{howard-ruder-2018-universal, peters2018deep, radford2019language} have led to a paradigm shift. One prominent architecture is BERT \cite{devlin2018bert}, a Transformer-based model that learns bidirectional encoder representations for words, on the basis of a masked language model and sentence adjacency training objective. Simply using BERT's representations in place of traditional embeddings has resulted in state-of-the-art performance on a range of downstream tasks including summarization \cite{liu2019bertsumm}, question answering and textual entailment \cite{devlin2018bert}. It is still, however, unclear why BERT representations perform well.

A flurry of recent work \cite{linzen2016assessing, gulordava-etal-2018-colorless, marvin2018targeted, lakretz2019emergence} has explored how recurrent neural language models perform in cases that require sensitivity to hierarchical syntactic structure, and study how they do so, particularly in the domain of agreement. In these studies, a pre-trained language model is asked to predict the next word in a sentence (a verb in the target sentence) following a sequence that may include other intervening nouns with different grammatical features (e.g., ``the \textbf{bear} by the \underline{trees} \textit{eats}..."). The predicted verb should agree with the subject noun (\textbf{bear}) and not the attractors (\underline{trees}), in spite of the latter's recency. Such analyses have revealed that LSTMs exhibit state tracking and explicit notions of word order for modeling long term dependencies, although this effect is diluted when sequential and structural information in a sentence conflict. Further work by \citet{gulordava-etal-2018-colorless} and others \cite{linzen2018distinct, giulianelli2018under} argues that RNNs acquire grammatical competence in agreement that is more abstract than word collocations, although language model performance that requires sensitivity to the phenomena such as reflexive anaphora, non-local agreement and negative polarity remains low \cite{marvin2018targeted}. 
Meanwhile, studies evaluating \emph{which} linguistic phenomena are encoded by contextualized representations \cite{goldberg2019assessing, wolf2019some, tenney2018you} successfully demonstrate that purely self-attentive architectures like BERT can capture hierarchy-sensitive, syntactic dependencies, and even support the extraction of dependency parses \cite{hewitt2019structural}. However, the way in which BERT does this has been less studied. In this paper, we investigate how and where the representations produced by pre-trained BERT models \cite{devlin2018bert} express the hierarchical organization of a sentence. 
 
We proceed in two ways. The first involves the use of diagnostic classifiers \cite{hupkes2018visual} to probe the presence of hierarchical and linear properties in the representations of words. However, unlike past work, we train these classifiers using a ``poverty of the stimulus" paradigm, where the training data admit both linear and hierarchical solutions that can be distinguished by an enriched generalization set.  This method allows us to identify what kinds of information are represented most robustly and transparently in the BERT embeddings.  We find that as we use embeddings from higher layers, the prevalence of linear/sequential information decreases, while the availability of on hierarchical information increases, suggesting that with each layer, BERT phases out positional information in favor of hierarchical features of increasing complexity.

In the second set of experiments, we explore a novel approach to the study of BERT's self-attention vectors. Past explorations of attention mechanisms, whether in the domain of vision \cite{olah2018the, carter2019activation} or NLP  \cite{bahdanau2014neural,karpathy2015visualizing,young2018recent,voita2018context}, have largely involved a range of visualization techniques or the study of the general distribution of attention.
Our work takes a quantitative approach to the study of attention and its encoding of syntactic dependencies. Specifically, we consider the relationships between verbs and the subjects with which they agree, and reflexive anaphors and their antecedents. Building on past work in psycholinguistics, we consider the influence of distractor noun phrases on the identification of these dependencies.  
We propose a simple attention-based metric called the {\em confusion score} that captures BERT's response to syntactic distortions in an input sentence. This score provides a novel \emph{quantitative} method of evaluating BERT's syntactic knowledge as encoded in its attention vectors. We find that BERT does indeed leverage syntactic relationships between words to preferentially attend to the ``correct" noun phrase for the purposes of agreement and anaphora, though syntactic structure does not show the strong categorical effects we sometimes find in natural language. This result again points to a representation of syntactically-relevant hierarchical information in BERT, this time through attention weightings.

Our analysis thus provides evidence that BERT's self-attention layers compose increasingly abstract representations of linguistic structure without explicit word order information, and that structural information is expressly favored over linear information. This explains why BERT can perform well on downstream NLP tasks, which typically require complex modeling of structural relationships.

\begin{table*}[ht!]
    \centering
    \resizebox{\textwidth}{!}{
    \begin{tabular}{ccc}
        \toprule
        & 
        \textbf{Main auxiliary task} & 
        \textbf{Subject noun task} \\
        \midrule
        \multirow{2}{*}{Training, Development} & the cat \underline{will} sleep & the \underline{bee} can sting \\
        & the cat \underline{will} eat the fish that can swim & the \underline{bee} can sting the boy \\
        \midrule
        \multirow{2}{*}{Generalization} & the cat that \textit{can} meow \underline{will} sleep & (compound noun) the \emph{queen} \underline{bee} can sting \\
        & the cat that \textit{can} meow \underline{will} eat the fish that can swim & (possessive) the \emph{queen}'s \underline{bee} can sting \\
        \bottomrule
    \end{tabular}
    }
    \caption{Representative sentences from the main auxiliary and subject noun tasks. For the latter, the generalization set contains two types of sentences, compound nouns and possessives, which are evaluated on separately. In each example, the correct token is underlined, while the distractor (consistent with the incorrect linear rule) is italicized.}
    \label{tb:dataset_classification}
\end{table*}

\section{Diagnostic Classification}\label{sec:diagclass}
For our first exploration of the kind of linguistic information captured in BERT's embeddings, we apply diagnostic classifiers to 3 tasks: identifying whether a given word is the sentence's \textbf{main auxiliary}, the sentence's \textbf{subject noun}, and the sentence's \textbf{$n^{th}$-token}. In each task, we assess how well BERT's embeddings encode information about a given linguistic property via the ability of a simple diagnostic classifier to correctly recover the presence of that property from the  embeddings of a single word. 
The three tasks focus on different sorts of information: identifying the main auxiliary and the subject noun requires sensitivity to  hierarchical or syntactic information, while the $n^{th}$-token requires linear information.

For each token in a given sentence, its input representation to BERT is a sum of its token, segment and positional embeddings \cite{devlin2018bert}. We refer to these inputs as \textbf{pre-embeddings}. Note that by construction, a) the pre-embeddings contain linear but not hierarchical information, and b) BERT cannot generate new linear information that is not already in the input. Thus, any linear information in BERT's embeddings ultimately stems from the pre-embeddings, while any hierarchical information must be constructed by BERT itself.

\subsection{Poverty of the stimulus}\label{sec:poverty}
To classify an embedding as a sentence's main auxiliary or subject noun, the network needs to have represented structural information about a word's role in the sentence. In many cases, such structural information can be approximated linearly: the main auxiliary or subject noun could be identified as the first auxiliary or noun in a sentence. Though such a linear generalization may be falsified if given certain complex examples, it will succeed over a large range of simple sentences. \citet{chomsky1980rules} argues that the relevant distinguishing examples may be very rare for the case of identifying the main auxiliary (a property that is necessary in order to form questions), and hence this is an instance of the ``poverty of the stimulus" that motivates the hypothesis of innate bias toward hierarchical generalizations. However, it seems clear that distinguishing examples are plentiful for the subject noun case. The question we are interested in, then, is whether and how BERT's embeddings, which result from training on a massive dataset, encode hierarchical information.

Pursuing the idea of poverty of the stimulus training \cite{mccoy2018revisiting}, we train diagnostic classifiers only on sentences in which the relevant property (main auxiliary or subject noun) is stateable in either hierarchical or sequential terms, i.e., the linearly first auxiliary or noun (cf.\ Section \ref{sec:class_data}). The classifiers are then tested on sentences of greater complexity in which the hierarchical and linear generalizations can be distinguished.  Since our classifier is a simple perceptron that can access only one embedding at a time, it cannot compute complex contingencies among the representations of multiple words, and cannot succeed unless such information is already encoded in the individual embeddings. Thus, success on these tasks would indicate that BERT robustly represents the words of a sentence using a feature space where the identification of hierarchical generalizations is easy.

\subsection{Dataset}\label{sec:class_data}
The main auxiliary and subject noun tasks use synthetic datasets generated from context-free grammars (cf.\ Appendix \ref{sec:apdx_cfg}) that were designed to isolate the relevant syntactic property for a poverty of the stimulus setup. Typical sentences are highlighted in Table \ref{tb:dataset_classification}. In both tasks, the training, development and generalization sets contained 40000, 10000, and 10000 examples respectively.

\paragraph{Main auxiliary}
In the training and development sets, the main auxiliary (\underline{will} in Table \ref{tb:dataset_classification}) is always the first auxiliary in the sentence. A classifier that learns the naive linear rule of identifying the first linearly occurring auxiliary instead of the correct hierarchical (syntactic) rule still performs well during training. However, in the generalization set, the subject of each sentence is modified by a relative clause that contains an intervening auxiliary (that \emph{can} meow). Since the main auxiliary is never the first auxiliary in this case, learning the hierarchical rule becomes imperative.

\paragraph{Subject noun}
In the training and development sets, the subject noun (\underline{bee} in Table \ref{tb:dataset_classification}) is always the first noun in the sentence. A classifier that learns the linear rule of identifying the first linearly occurring noun does well during training, but only the hierarchical rule gives the right answer at test time. In the generalization set (both compound nouns \& possessives cases), the subject noun is the head of the construction (\underline{bee}) and not the dependent (\emph{queen}). In the possessives case, we note that subword tokenization always produces {\em 's} as a standalone token, e.g. {\em queen's} is tokenized into {\em [queen]} {\em ['s]}. Also, we allow sentences to chain an arbitrary number of possessives via nesting.

\paragraph{$n^{th}$-token}
For this experiment, we use sentences from the Penn Treebank WSJ corpus. Following the setup of \citet{collins-2002-discriminative} and filtering for sentences between 10 to 30 tokens BERT tokenization, we obtained training, development and generalization sets of sentences of sizes 21142, 3017 and 2999. We only consider $2 \leq n \leq 9$. In particular, we ignore $n = 1$ since the first token produced by BERT is always trivially [CLS].

\subsection{Methods}\label{sec:class_methods}
\paragraph{BERT models}
In our experiments, we consider two of Google AI's pre-trained BERT models \emph{bert-base-uncased} (\textbf{bbu}) and \emph{bert-large-uncased} (\textbf{blu}) from a PyTorch implementation.\footnote{\url{https://github.com/huggingface/pytorch-pretrained-BERT}} bbu has 12 layers, 12 attention heads and embedding width 768, while blu has 24 layers, 16 attention heads and embedding width 1024.

\paragraph{Training}
For each task, we train a simple perceptron with a sigmoid output to perform binary classification on individual token embeddings of a sentence, based on whether the underlying token possesses the property relevant to the task. This is similar to the concept of diagnostic classification by \citet{hupkes2018visual, giulianelli2018under}. 

\begin{table}[h!]
\centering
\resizebox{0.65\columnwidth}{!}{
\begin{tabular}{ccccc}
 & the & cat & will & sleep \\
(BERT) & $\downarrow$ & $\downarrow$ & $\downarrow$ & $\downarrow$ \\
 & $e_1$ & $e_2$ & $e_3$ & $e_4$ \\
(Classifier) & $\downarrow$ & $\downarrow$ & $\downarrow$ & $\downarrow$ \\
 & $\widehat{y}_1$ & $\widehat{y}_2$ & $\widehat{y}_3$ & $\widehat{y}_4$ \\
\end{tabular}
}
\end{table}

Each input sentence is tokenized and processed by BERT, and the resulting embeddings $\{e_i\}$ are individually passed to the classifier $f_\theta$ to produce a sequence of logits $\{\widehat{y}_i\}$.
Supervision is provided via a one-hot vector of indicators $\{y_i\}$ for the specified property. For example, in the main auxiliary task, the above example would have $y_1 = y_2 = y_4 = 0$ and $y_3 = 1$, since the third word is the main auxiliary. The contribution of each example to the total cross-entropy loss is:
\begin{equation}\label{eqn:xeloss}
    \mathcal{L}_\theta = -\sum_i (y_i \log \widehat{y}_i + (1-y_i) \log (1-\widehat{y}_i) )
\end{equation}

Each classifier is trained for a single epoch using the Adam optimizer \cite{kingma2014adam} with hyperparameters $lr = 0.001, \beta_1 = 0.9, \beta_2 = 0.999$. We freeze BERT's weights throughout training, which allows us to take good classification performance as evidence that the information relevant to the task is being encoded in BERT's embeddings in a salient, easily retrievable manner.

\paragraph{Evaluation}
For each example at test time, after computing the logits we obtain the index of the classifier's most confident guess within the sentence:
\begin{equation}
    i^* = \argmax_i \widehat{y_i}
\end{equation}
The average $y_{i^*}$ across the test set is reported as the classification accuracy.

\paragraph{Layerwise diagnosis}
One key aspect of our experiments is the training of layer-specific classifiers for \emph{all} layers. This yields a \emph{layerwise} diagnosis of the information content in BERT's embeddings, providing a glimpse into how BERT internally manipulates and composes linguistic information. We also train classifiers on the pre-embeddings, which can be considered as the ``zero-th" layer of BERT and hence act as useful baselines for content present in the input.

\subsection{Results}
\begin{figure}[!t]
    \centering
    \includegraphics[width=0.48\textwidth]{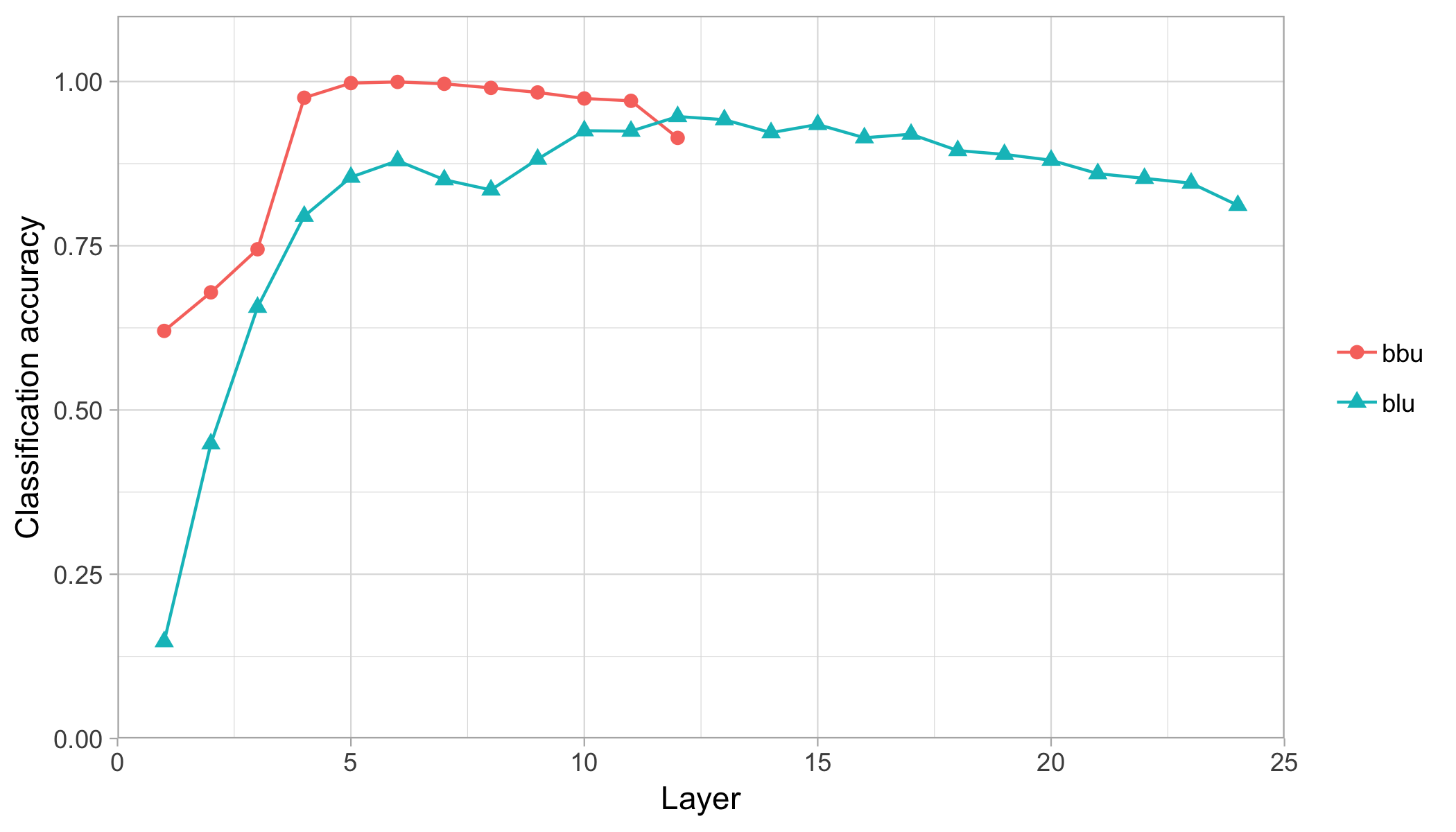}
    \vspace{-5mm}
    \caption{Layerwise accuracy of diagnostic classifiers on the generalization set of the main auxiliary task.}
\label{fig:main_aux_gen}
\vspace{-2mm}
\end{figure}

\paragraph{Main auxiliary}
Classifiers for both models achieved near-perfect accuracy across all layers on the development set. In Figure \ref{fig:main_aux_gen}, we observe that on the generalization set, the classifiers for both models can identify the main auxiliary with over 85\% accuracy past layer 5, and bbu in particular obtains near-perfect accuracy from layers 4 to 11.

As discussed in Section \ref{sec:class_data}, the classifiers were only given training examples where the main auxiliary was also the first auxiliary in the sentence. 
Although the linear rule ``pick the first auxiliary" is compatible with the training data, the classifier nonetheless learns the more complex but correct hierarchical rule ``pick the auxiliary of the main clause". By our argument from Section \ref{sec:poverty}, this suggests that BERT embeddings encode syntactic information relevant to whether a token is the main auxiliary, as a feature salient enough to be recoverable by our simple diagnostic classifier.

We found that almost all instances of classification errors involved the misidentification of the linearly first auxiliary (within the relative clause) as the main auxiliary, e.g. \emph{can} instead of \underline{will} in Table \ref{tb:dataset_classification}. We believe that this stems from the significance
of part-of-speech information for language modeling.  As a result, any word of a different POS will not be chosen by the classifier.

\begin{figure}[!t]
    \centering
    \includegraphics[width=0.5\textwidth]{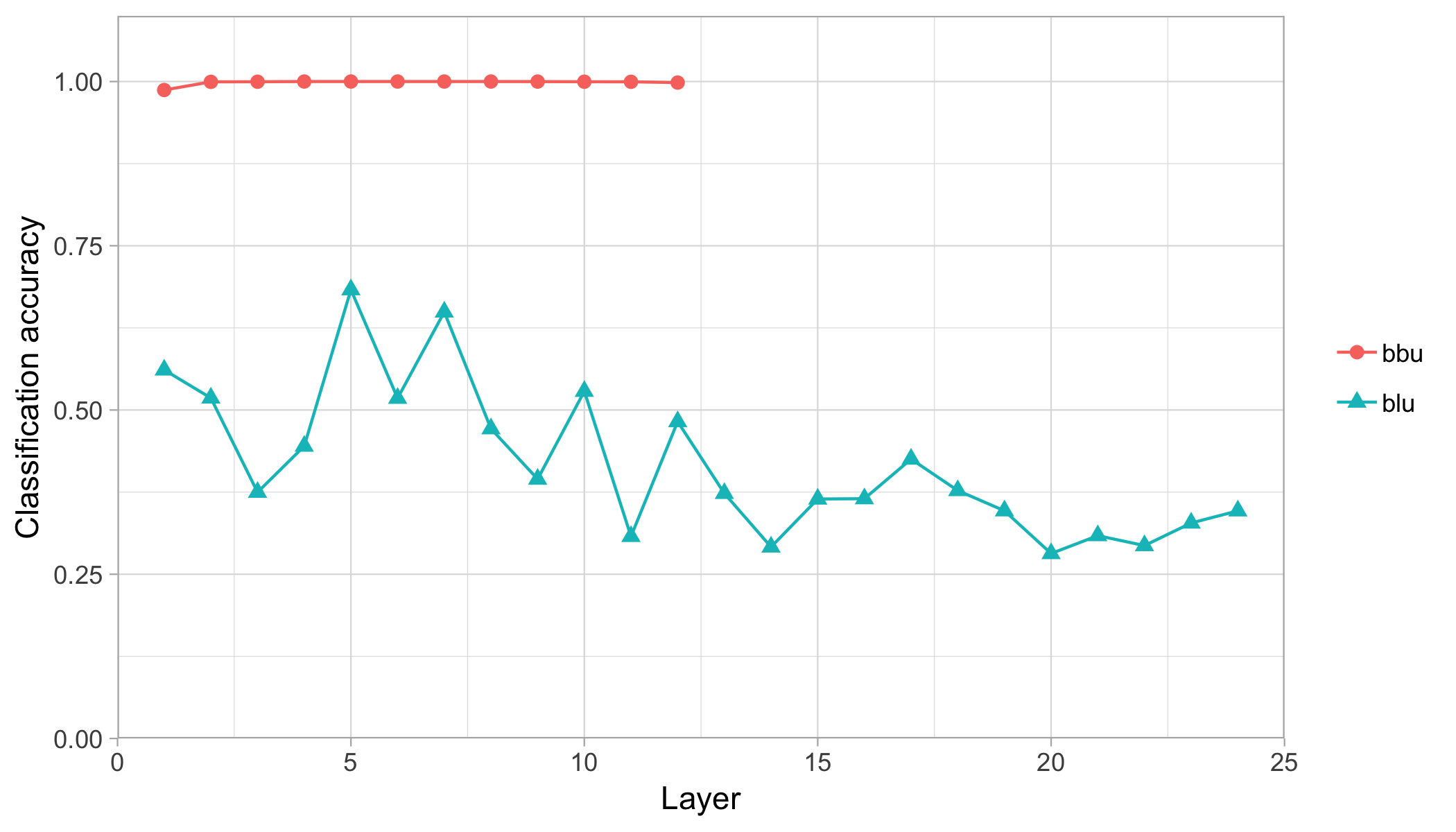}
    \vspace{-5mm}
    \caption{Layerwise accuracy of diagnostic classifiers on the compound noun generalization set of the subject noun task.}
\label{fig:sbjn_cpd_gen}
\vspace{-2mm}
\end{figure}

\begin{figure}[!t]
    \centering
    \includegraphics[width=0.5\textwidth]{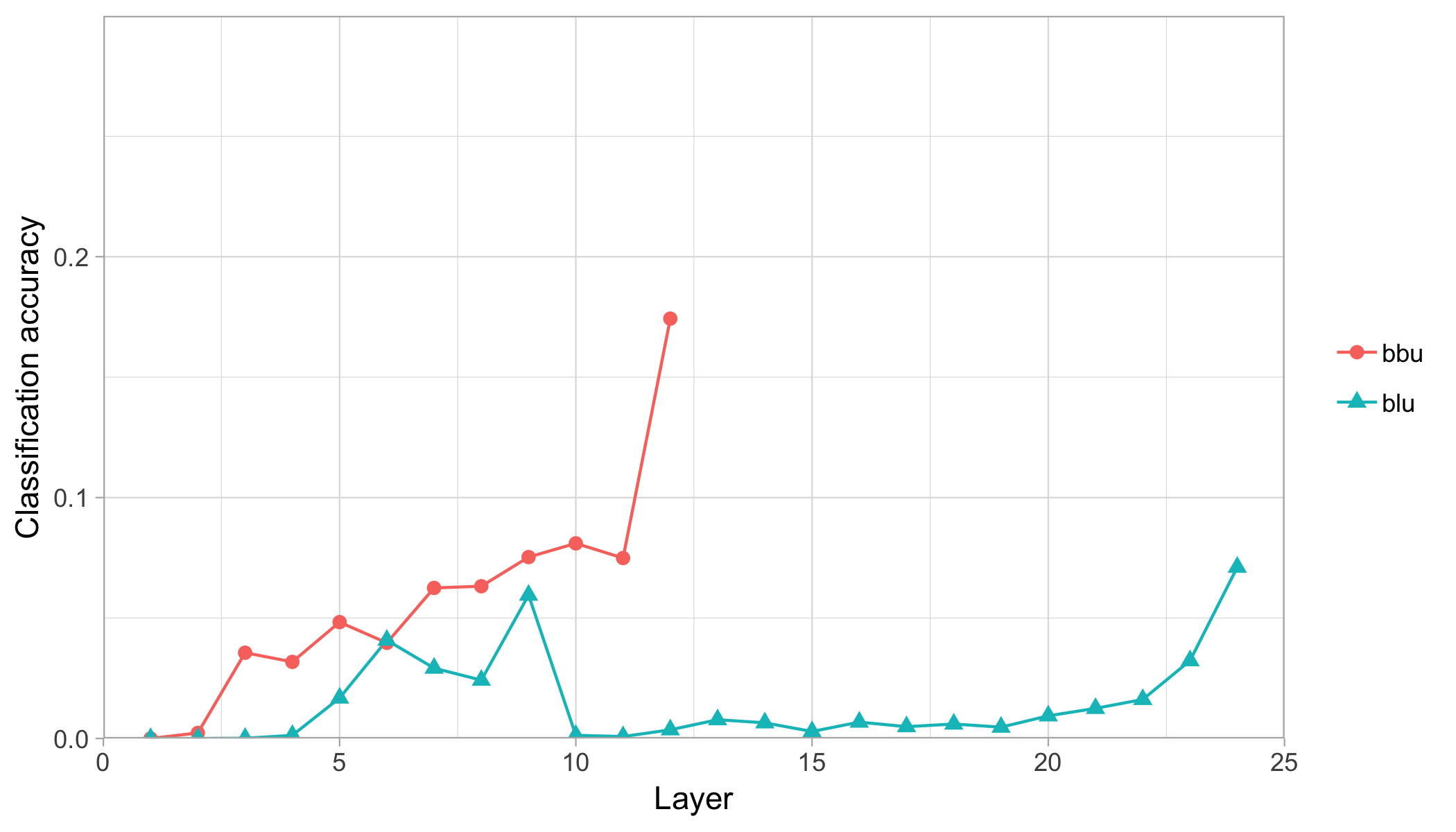}
    \vspace{-5mm}
    \caption{Layerwise accuracy of diagnostic classifiers on the possessive generalization set of the subject noun task.}
\label{fig:sbjn_poss_gen}
\vspace{-2mm}
\end{figure}

\paragraph{Subject noun}
As with the previous task, classifiers for both models achieved near-perfect accuracy across all layers on the development set. On the compound noun generalization set (Figure \ref{fig:sbjn_cpd_gen}), while bbu achieved near-perfect accuracy in later layers, blu consistently performed poorly. bbu's performance suggests that in the classifier successfully learns a generalization that excludes the first noun of a compound, as opposed to the naive linear rule ``pick the first noun". As before, this suggests that BERT encodes syntactic information in its embeddings. However, blu's performance is unexpected: it consistently predicts the object noun when it makes errors. In contrast, on the possessive generalization set (Figure \ref{fig:sbjn_poss_gen}), both models perform poorly. We offer an explanation for this distinctive performance in Section \ref{sec:class_further}.

\begin{figure}[!t]
    \centering
    \includegraphics[width=0.5\textwidth]{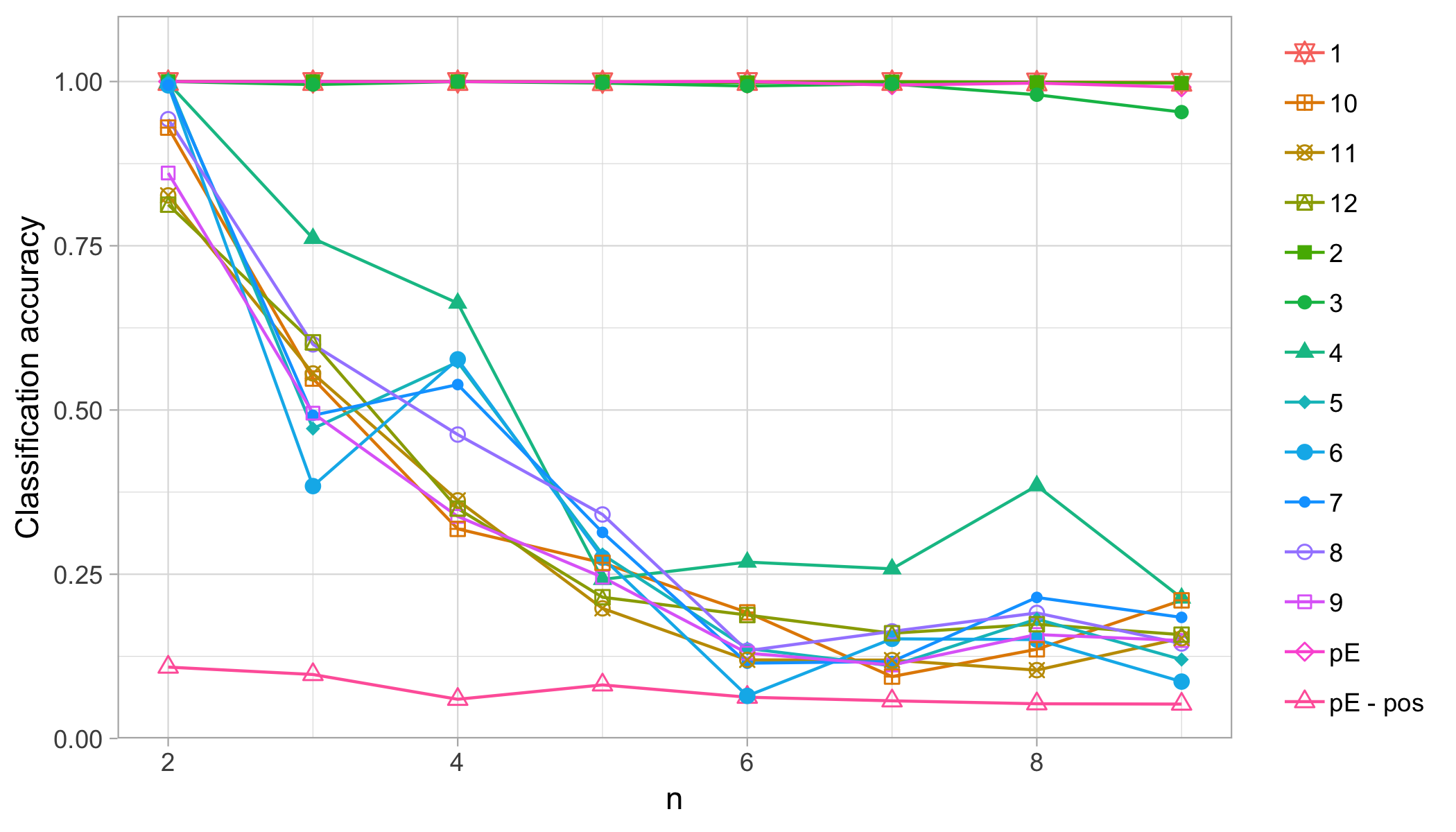}
    \vspace{-5mm}
    \caption{Layerwise accuracy of diagnostic classifiers on the generalization set of the $n^{th}$ token task, for the bbu model only. Note that each line corresponds to a particular layer's embeddings as we vary $2 \leq n \leq 9$. pE denotes pre-embeddings and pE -- pos denotes pre-embeddings without the positional component.}
    \label{fig:nth_token_gen}
\vspace{-2mm}
\end{figure}

\paragraph{$n^{th}$ token}
Since this property is entirely determined by the linear position of a word in a sentence, it directly measures the amount of positional information encoded in the embeddings. Here we have two baselines characterizing both extremes: the normal pre-embeddings (denoted pE) and a variant (pE -- pos) where we exclude the positional component from its construction. Since BERT cannot introduce any new positional information, we expect these two to represent upper and lower bounds on the amount of positional information present in BERT's embeddings.

In Figure \ref{fig:nth_token_gen}, we see a dramatic difference in performance on pE (one of the topmost lines) compared to pE -- pos (bottommost line). We note that performances across all 12 layers fall between these two extremes, confirming our intuitions from earlier. Specifically, the classifiers for layers 1 -- 3 have near-perfect accuracy on identifying an arbitrary $n^{th}$ token ($2 \leq n \leq 9$). However, from layer 4 onwards, the accuracy drops sharply as $n$ increases. This suggests that the positional component of pre-embeddings is the primary source of positional information in BERT, and BERT (bbu) discards a significant amount of positional information between layers 3 and 4, possibly in favor of hierarchical information.

\subsection{Further Analysis} \label{sec:class_further}

\paragraph{Main auxiliary}
In Figure \ref{fig:main_aux_gen}, we observe that classification accuracy increases sharply in the first 4 layers, then plateaus before slowly decreasing. This mirrors a similar layerwise trend observed by \citet{hewitt2019structural}.
We postulate that the embeddings reach their ``optimal level of abstraction" with respect to their ability to predict the main auxiliary halfway through BERT (about layer 6 for bbu, 12 for blu). At layer 1, the embedding for each token is a highly localized representation that contains insufficient sentential context to determine whether it is the main auxiliary of the sentence. As layer depth increases, BERT composes increasingly abstract representations via self-attention, which allows it to extract information from other tokens in the sentence. At some point, the representation becomes abstract enough to represent the hierarchical concept of a ``main auxiliary", causing an early increase in classification accuracy. However, as depth increases further, the representations become so abstract that finer linguistic features are increasingly difficult to recover, e.g., a token embedding at the sentence-vector level of abstraction may longer be capable of identifying itself as the main auxiliary, accounting for the slowly deteriorating performance towards later layers.

\paragraph{Subject noun}
Given the similarity of the main auxiliary and the subject noun classification tasks, we might expect them to exhibit similar trends in performance. In Figure \ref{fig:sbjn_cpd_gen}, we observe a similar early increase in diagnostic classification accuracy for the bbu embeddings. The lack of significant performance decay on higher layers possibly reflects the salience of the subject noun feature even at the sentence-vector level of abstraction.
Strangely, blu performed poorly, even worse than chance (50\%). We are unable to explain why this happens and leave this for future research.

On the possessive generalization set, the poor performance of both models seems to contradict the hypothesis that BERT has learned an abstract hierarchical generalization to classify subject nouns. We conjecture that BERT's issues in the possessive case stem from the ambiguity of the {\em 's} token, which can function either as a possessive marker or as a contracted auxiliary verb (e.g.``She's sleeping"). If BERT takes a possessive occurrence of {\em 's} as the auxiliary verb, the immediately preceding noun can be (incorrectly) analyzed as the subject. If so, this would suggest that BERT does not represent the syntactic structure of the entire sentence in a unified fashion, but instead uses local cues to constituency. In Figure \ref{fig:sbjn_poss_gen}, the gradually increasing but still poor performance towards later layers in both models suggests that the embeddings might be trending toward a more abstract representation, but do not ultimately achieve it.

\paragraph{$n^{th}$ token}
For each layer $k \geq 3$, Figure \ref{fig:nth_token_gen} shows an asymmetry where the classifier for layer $k$ performs worse at identifying the $n^{th}$ token as $n$ increases. We believe that this may be an artifact of the distributional properties of natural language: the distribution of words that occur at the start of a sentence tends to be concentrated on a small class of parts of speech that can occur near the beginning of constituents that can begin a sentence. As $n$ increases, the class of possible parts is no longer a function of the beginning of the sentence, and as a result becomes more uniform. As a result, it is easier for a classifier to predict whether a given word is the $n^{th}$ token when $n$ is small, since it can make use of easily accessible part-of-speech information in the embeddings to limit its options to only the tokens likely to occur in a given position.
\begin{table*}[ht!]
    \centering
    \resizebox{\textwidth}{!}{
    \begin{tabular}{cccclc}
        \toprule
         \multicolumn{6}{c}{ Subject-Verb Agreement } \\
        \midrule
        \textbf{Condition} & 
        \textbf{\thead{Relative \\ Clause}} & 
        \textbf{\thead{DN \\ Number Match}} & &
        \textbf{Example Sentence} &
        \textbf{\thead{Mean \\ Confusion Score}} \\
        \midrule
        A1 & \xmark & \checkmark & & 
        \underline{the cat} near \emph{the dog} \underline{does} sleep & 0.97 \\
        A2 & \xmark & \xmark & &
        \underline{the cat} near \emph{the dogs} \underline{does} sleep & 0.93 \\
        A3 & \checkmark & \checkmark & &
        \underline{the cat} that can comfort \emph{the dog} \underline{does} sleep & 0.85 \\
        A4 & \checkmark & \xmark & &
        \underline{the cat} that can comfort \emph{the dogs} \underline{does} sleep & 0.81 \\
        \midrule
        \multicolumn{6}{c}{ Reflexive Anaphora } \\
        \midrule
        \textbf{Condition} & 
        \textbf{\thead{Relative \\ Clause}} & 
        \textbf{\thead{DN\textsubscript{o} \\ Gender Match}} & 
        \textbf{\thead{DN\textsubscript{r} \\ Gender Match}} & 
        \textbf{Example Sentence} &
        \textbf{\thead{Mean \\  Confusion Score}} \\
        \midrule
        R1 & \xmark & \checkmark & NA & \underline{the lord} could comfort \emph{the wizard} by \underline{himself} & 1.01 \\
        R2 & \xmark & \xmark & NA & \underline{the lord} could comfort \emph{the witch} by \underline{himself} & 0.92 \\
        R3 & \checkmark & NA & \checkmark & \underline{the lord} that can hurt \emph{the prince} could comfort \underline{himself} & 0.99 \\
        R4 & \checkmark & NA & \xmark & \underline{the lord} that can hurt \emph{the princess} could comfort \underline{himself} & 0.89 \\
        R5 & \checkmark & \checkmark & \checkmark & \underline{the lord} that can hurt \emph{the prince} could comfort \emph{the wizard} by \underline{himself} & 1.57 \\
        R6 & \checkmark & \checkmark & \xmark & \underline{the lord} that can hurt \emph{the princess} could comfort \emph{the wizard} by \underline{himself} & 1.52 \\
        R7 & \checkmark & \xmark & \checkmark & \underline{the lord} that can hurt \emph{the prince} could comfort \emph{the witch} by \underline{himself} & 1.49 \\
        R8 & \checkmark & \xmark & \xmark & \underline{the lord} that can hurt \emph{the princess} could comfort \emph{the witch} by \underline{himself} & 1.39 \\
        \bottomrule
    \end{tabular}
    }
    \caption{Representative sentences from the subject-verb agreement and reflexive anaphora datasets for each condition, and corresponding mean confusion scores. DN\textsubscript{o}: distractor noun as object. DN\textsubscript{r}: distractor noun in relative clause.}
    \label{tb:dataset_result_attention}
\end{table*}

\section{Diagnostic Attention}
Our second exploration of BERT's syntactic knowledge focuses on the encoding of grammatical relationships instead of the identification of  elements with specific structural properties. We consider two phenomena: {\bf reflexive anaphora} and {\bf subject-verb agreement}. For each, we determine the extent to which BERT attends to linguistically relevant elements via the self-attention mechanism. This gives us further information about \emph{how} hierarchy-sensitive syntactic information is encoded.

\subsection{Quantifying intrusion effects via attention}

Subject-verb and antecedent-anaphor dependencies both involve a dependent element, which we call the {\em target} (the verb or the anaphor) and the element on which it depends, which we call the {\em trigger} (the subject or the antecedent that provides the interpretation). A considerable body of work in psycholinguistics has explored how humans process such dependencies in the presence of elements that are not in relevant structural positions but which linearly intervene between the trigger and target. \citet{dillon2013contrasting} aim to quantify this intrusion effect in human reading for the two dependencies we explore here. Under the assumption that higher reading time and eye movement regressions indicate an intrusion effect, they conclude that intruding noun-phrases have a substantial effect on the processing of subject-verb agreement, but not antecedent-anaphor relations.

We adapt this idea in measuring intrusion effects in BERT. We propose a simple and novel metric we term the ``confusion score" for quantifying intrusion effects using attention. This quantitative metric allows us to measure the preferable attention of transformer-based self-attention on one entity as opposed to another. Formally, suppose $X = \{x_i\}_{i=1}^n$ are linguistic units of interest, i.e. candidate triggers for the dependency, and $Y$ is the dependency target. For each layer $l$ and attention head $a$, we sum the self-attention weights from the indices of $x_i$ (since each $x_i$ may consist of multiple words) on attention head $a$ of layer $l-1$ to $Y$ on layer $l$, and take the mean over $A$ attention heads:
\begin{equation}
    \text{attn}_l(x_i,Y) = \frac{1}{A} \sum_{a=1}^A \sum_{x_{ij} \in x_i} \text{attn}_{la}(x_{ij},Y)
\end{equation}
We finally define the {\bf confusion score} on layer $l$ as the binary cross entropy of the normalized attention distribution between $\{x_i\}$ given $Y$ as follows:  
\begin{equation}\label{eqn:conf}
    \text{conf}_l(X,Y) = - \log \frac{\text{attn}_l(x_1,Y)}{\sum_{i=1}^n \text{attn}_l(x_i,Y)}
\end{equation}
Note that this equation assumes that each dependency has a unique trigger $x_1$: verbs agree with a single subject, and anaphors take a single noun phrase as their antecedent. 

Our study focuses on the examples of the forms shown in Table \ref{tb:dataset_result_attention}. For subject-verb agreement, there are two types of examples: with the distractor within a PP (A1 and A2) and with the distractor within a RC (A3 and A4). Past psycholinguistic work has shown that distractor noun phrases within PPs give rise to greater processing difficulty than distractors within RCs \cite{bock1992regulating}. For each type, we compare confusion in the case of distractors that share features with the subject, the true trigger of agreement, (A1 and A3) with those that do not (A2 and A4). Our expectation is that distractors that do not share features with the target of agreement will yield less confusion.

For reflexive anaphora, because of the possibility of ambiguity, we also consider sentences that include a noun phrase that is a structurally possible antecedent. For example, condition R1 has the subject {\em the lord} as its antecedent, but the object noun phrase {\em the wizard} is also grammatically possible. In contrast, for R2, the mismatch in gender features prevents the object from serving as an antecedent, which should lead to lower confusion. Sentences R3 and R4 include a distractor noun phrase within a RC. Since this noun phrase does not c-command the anaphor, it is grammatically inaccessible and should therefore contribute less, if at all, to confusion. Sentence types R5 through R8 include both the relative modifier and the object noun phrase, and systematically vary the agreement properties of the two distractors.  

We hypothesize that attention weights on each linguistic unit indicate the relative importance of that entity as a trigger of a linguistic dependency. As a result, the ideal attention distribution should put all of the probability mass on the antecedent noun phrase for reflexive anaphora or on the subject noun phrase for agreement, and zero on the distractor noun phrases. As a baseline, a uniform distribution over two noun phrases, one the actual target and the other a distractor, would lead to a confusion score of  $-\log \frac{1}{2} = 1$; with two distractors, the uniform probability baseline would be $-\log \frac{1}{3} = 1.6$.

\subsection{Dataset}
We construct synthetic datasets using context-free grammars (shown in Appendix \ref{sec:apdx_cfg}) for both  subject-verb agreement and reflexive anaphora and compute mean confusion scores across multiple sentences. This allows us to control for semantic effects on the confusion score. All datasets for each condition contain 10000 examples. 

In the subject-verb agreement datasets, we vary 1) the type of subordinate clause (prepositional phrase, PP; or relative clause, RC), and 2) the number on the distractor noun phrase. All conditions should be unambiguous, since only the head noun phrase can agree with the auxiliary. 

In the reflexive anaphora datasets, we vary 1) the presence of a RC, 2) the gender match between the RC's noun phrase and the reflexive 3) the presence of an object noun phrase, and 4) the gender match between the object noun and the reflexive. All nouns are singular. Conditions R1, R5, R6 are ambiguous conditions, as they include an object noun phrase that matches the reflexive in gender. In other conditions, only the head noun phrase is the possible antecedent: the object mismatches in features and the noun phrase within the RC is grammatically inaccessible.

\subsection{Methods}
We use Equation \ref{eqn:conf} to compute the confusion score on each layer for the target in each sentence in our dataset. As in Section \ref{sec:class_methods}, this yields a \emph{layerwise} diagnosis of confusion in BERT's self-attention mechanism. We also compute the mean confusion score across all layers.\footnote{We built on \citet{vig2019visualizing}'s BERT attention visualization library \url{https://github.com/jessevig/bertviz} to implement the attention-based confusion score.} In our experiments, we compute confusion scores using bbu only.

Note that in conditions R1, R5 and R6, there are two possible antecedents of the reflexive. We nonetheless use Equation \ref{eqn:conf} to calculate confusion scores relative to a single antecedent (the subject). 

To compute the significance of the presence of different types of distractors and of feature mismatch of the distractors, we run a linear regression to predict confusion score. For subject-verb agreement, the baseline value is the confusion at layer 1 of a sentence with a PP and a mismatch in number on the distractor noun (condition A2 in Table \ref{tb:dataset_result_attention}). For reflexive anaphora, the baseline is the confusion at layer 1 of a sentence with no RC and no object noun (e.g. ``the lord comforts himself").

\subsection{Results}
\begin{figure}[!t]
    \centering
    \includegraphics[width=0.5\textwidth]{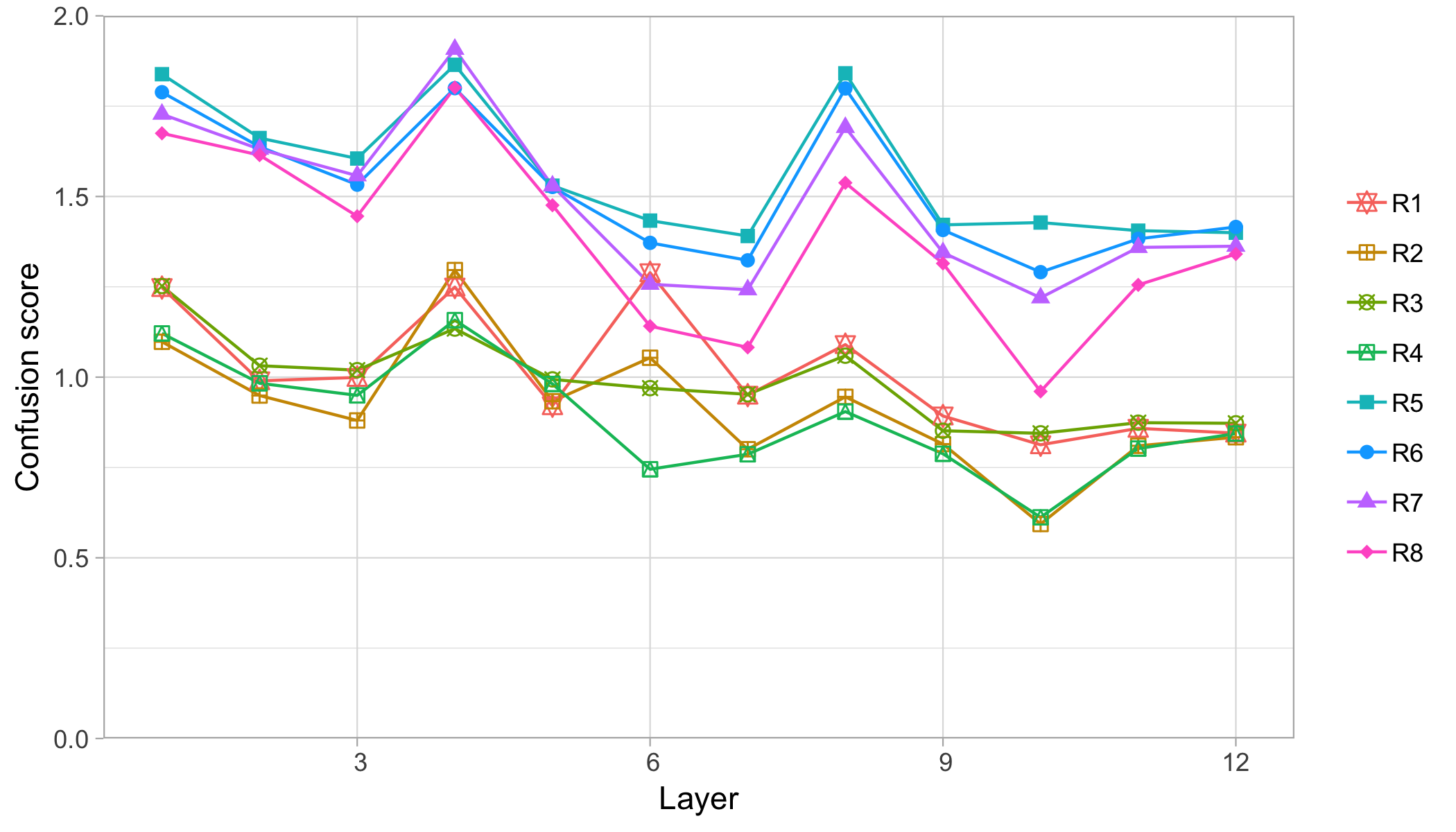}
    \vspace{-5mm}
    \caption{Layerwise confusion scores for each reflexive anaphora condition listed in Table \ref{tb:dataset_result_attention}. Conditions R1 to R4 have one distractor noun phrase, but conditions R5 to R8 have two distractor noun phrases.}
    \label{fig:attn_refl}
\vspace{-2mm}
\end{figure}


\begin{table}[t]
    \centering
    \resizebox{\columnwidth}{!}{
    \begin{tabular}{lrc}
        \toprule
        \textbf{Coefficient} & 
        \multicolumn{1}{c}{\textbf{Estimate}} & 
        \textbf{p-value} \\
        \midrule
         \multicolumn{3}{c}{ Subject-Verb Agreement } \\
        \midrule
        Intercept & $1.33 \pm 1.32 \mathrm{e}{-3}$ & $< 2\mathrm{e}{-16}$ \\
        Relative Clause & $-0.12 \pm 1.03 \mathrm{e}{-3}$ & $< 2\mathrm{e}{-16}$ \\
        DN\textsubscript{r} Number Match & $0.03 \pm 1.03 \mathrm{e}{-3}$ & $< 2\mathrm{e}{-16}$ \\
        Layer & $-0.06 \pm 1.50 \mathrm{e}{-4}$ & $< 2\mathrm{e}{-16}$ \\
        \midrule
       \multicolumn{3}{c}{ Reflexive Anaphora } \\
        \midrule
        Intercept & $0.63 \pm 1.24 \mathrm{e}{-3}$ & $< 2\mathrm{e}{-16}$ \\
        DN\textsubscript{o} Gender Match & $ 0.60 \pm 9.09 \mathrm{e}{-4}$ & $< 2\mathrm{e}{-16}$ \\
        DN\textsubscript{o} Gender Mismatch & $ 0.50 \pm 9.09 \mathrm{e}{-4}$ & $< 2\mathrm{e}{-16}$ \\
        DN\textsubscript{r} Gender Match & $ 0.57 \pm 9.09 \mathrm{e}{-4}$ & $< 2\mathrm{e}{-16}$ \\
        DN\textsubscript{r} Gender Mismatch & $ 0.49 \pm 9.09 \mathrm{e}{-4}$ & $< 2\mathrm{e}{-16}$ \\
        Layer & $-0.03 \pm 9.72 \mathrm{e}{-5}$ & $< 2\mathrm{e}{-16}$ \\
        \bottomrule
    \end{tabular}
    }
    \caption{Regression estimates and p-values for the coefficient effects under reflexive anaphora and subject-verb agreement. All effects are statistically significant.}
    \label{tb:linreg_attn}
\end{table}

\paragraph{Subject-verb agreement}
Since sentence types A1 to A4 are all unambiguous, ideal confusion scores should be zero. However, Table \ref{tb:dataset_result_attention} indicates that the mean confusion scores are instead closer to the uniform probability baseline confusion score of $1$, suggesting that BERT's self-attention mechanism is far from able to perfectly model syntactically-sensitive hierarchical information. Nonetheless, from Table \ref{tb:linreg_attn}, we see that BERT's attention mechanism is in fact sensitive to subtleties of linguistic structure: a distractor within a PP causes more confusion than one within a relative clause (i.e., the presence of the relative has a negative coefficient in the linear model), in agreement with past psycholinguistic work \cite{bock1992regulating}. Moreover, the presence of matching distractors has a significant positive effect on confusion scores. These findings therefore suggest that BERT representations are sensitive to different types of syntactic embedding as well as the values of number features in computing subject-verb agreement dependencies.

\paragraph{Reflexive anaphora}
From Table \ref{tb:dataset_result_attention}, we see the major effect of the number of distractor noun phrases: mean confusion scores for conditions with one distractor (R1-R4) are lower than those with two distractors (R5-R8). If BERT were perfectly exploiting grammatical structure, we should expect the presence of a grammatically inaccessible distractor noun within a relative clause not to add to confusion. Thus, we might expect R5 and R6 to have mean confusion scores comparable to R1, as both include single grammatically viable distractor. However, they both have higher mean confusion scores than R1 (the same is true for R7/R8 vs.\ R2). Moreover, conditions R2 to R4 and R7 to R8 should have confusion scores of zero, since the head noun phrase is the only grammatically possible antecedent. This, however, is not so. Taken together, we might conclude that BERT attends unnecessarily to grammatically inaccessible or grammatically mismatched distractor noun phrases, suggesting that it does not accurately model reflexive dependencies. 

Nonetheless, if we look more closely at the effects of the different factors through the linear model reported in Table \ref{tb:linreg_attn}, we once again find evidence for a sensitivity to both syntactic structure and grammatical features: the presence of grammatically accessible distractors has a (slightly) larger effect on confusion than grammatically inaccessible distractors (i.e., DN\textsubscript{o} vs.\ DN\textsubscript{r}), particularly when the distractor matches in features with the actual antecedent. 


\subsection{Further Analysis}
\paragraph{Layerwise diagnosis}
Figure \ref{fig:attn_refl} and Table \ref{tb:linreg_attn} show that confusion is negatively correlated with layer depth for reflexive anaphora. Confusion scores for subject-verb agreement exhibit a similar trend. This provides additional evidence for our conjecture that BERT composes increasingly abstract representations containing hierarchical information, with an optimal level of abstraction. Notably, the observed sensitivity of BERT's self-attention values to grammatical distortions suggests that BERT's syntactic knowledge is in fact encoded in its attention matrices. Finally, it is worth noting that confusion for both reflexives and subject-verb agreement showed an increase at layer 4. Strikingly, this was the level at which linear information was found, through diagnostic classifiers, to be degraded. We leave for the future an understanding of the connection between these.

\section{Conclusion}
In this paper, we investigated how and to what extent BERT representations encode syntactically-sensitive hierarchical information, as opposed to linear information. Through diagnostic classification, we find that positional information is encoded in BERT from the pre-embedding level up through lower layers of the model. At higher layers, information becomes less positional and more hierarchical, and BERT encodes increasingly complex representations of sentence units. 

We propose a simple and novel method of observing, for a given syntactic phenomenon, the intrusion effects of distractors on BERT's self-attention mechanism. Through such diagnostic attention, we find that BERT does encode aspects of syntactic structure that are relevant for subject-verb agreement and reflexive dependencies through attention weights, and that this information is represented more accurately on higher layers. We also find evidence that BERT is responsive to matching of grammatical features such as gender and number. 
However, BERT's attention is only incompletely modulated by structural and featural properties, and attention is sometimes spread across grammatically irrelevant elements.

We conclude that BERT composes increasingly abstract hierarchical representations of linguistic structure using its self-attention mechanism. To further understand BERT's syntactic knowledge, further work can be done to (1) investigate or visualize layer-on-layer changes in BERT's structural and positional information, particularly between layers 3 and 4 when positional information is largely phased out, and (2) retrieve the increasingly hierarchical representations of BERT across layers via the self-attention mechanism.

\bibliography{acl2019}
\bibliographystyle{acl_natbib}

\appendix
\vfill\clearpage
\section{Appendix}
\subsection{Context-free grammars for dataset generation}\label{sec:apdx_cfg}

\begin{figure}[H]
    \centering
    \resizebox{0.95\textwidth}{!}{
    \begin{tabular}{lcl}
        \hline
        \thead[l]{S} & $\rightarrow$ & \thead[l]{NP\textsubscript{M} VP\textsubscript{M}} \\
        \thead[l]{NP\textsubscript{M}} & $\rightarrow$ & \thead[l]{Det N $\mid$ Det N Prep Det Nom $\mid$ Det N RC} \\
        \thead[l]{NP\textsubscript{O}} & $\rightarrow$ & \thead[l]{Det Nom $\mid$ Det Nom Prep Det Nom $\mid$ Det Nom RC} \\
        \thead[l]{VP\textsubscript{M}} & $\rightarrow$ & \thead[l]{Aux VI $\mid$ Aux VT NP\textsubscript{O}} \\
        \thead[l]{RC} & $\rightarrow$ & \thead[l]{Rel Aux VI $\mid$ Rel Det Nom Aux VT $\mid$ Rel Aux VT Det Nom} \\
        \thead[l]{Nom} & $\rightarrow$ & \thead[l]{N $\mid$ JJ Nom} \\
        \thead[l]{Det} & $\rightarrow$ & \thead[l]{the $\mid$ some $\mid$ my $\mid$ your $\mid$ our $\mid$ her} \\
        \thead[l]{N} & $\rightarrow$ & \thead[l]{bird $\mid$ bee $\mid$ ant $\mid$ duck $\mid$ lion $\mid$ dog $\mid$ tiger $\mid$ worm $\mid$ horse $\mid$ cat $\mid$ fish $\mid$ bear $\mid$ wolf $\mid$ birds $\mid$ bees $\mid$ ants $\mid$ \\ ducks $\mid$ lions $\mid$ dogs $\mid$ tigers $\mid$ worms $\mid$ horses $\mid$ cats $\mid$ fish $\mid$ bears $\mid$ wolves} \\
        \thead[l]{VI} & $\rightarrow$ & \thead[l]{cry $\mid$ smile $\mid$ sleep $\mid$ swim $\mid$ wait $\mid$ move $\mid$ change $\mid$ read $\mid$ eat} \\
        \thead[l]{VT} & $\rightarrow$ & \thead[l]{dress $\mid$ kick $\mid$ hit $\mid$ hurt $\mid$ clean $\mid$ love $\mid$ accept $\mid$ remember $\mid$ comfort} \\ 
        \thead[l]{Aux} & $\rightarrow$ & \thead[l]{can $\mid$ will $\mid$ would $\mid$ could} \\
        \thead[l]{Prep} & $\rightarrow$ & \thead[l]{around $\mid$ near $\mid$ with $\mid$ upon $\mid$ by $\mid$ behind $\mid$ above $\mid$ below} \\
        \thead[l]{Rel} & $\rightarrow$ &\thead[l]{who $\mid$ that} \\
        \thead[l]{JJ} & $\rightarrow$ & \thead[l]{small $\mid$ little $\mid$ big $\mid$ hot $\mid$ cold $\mid$ good $\mid$ bad $\mid$ new $\mid$ old $\mid$ young} \\\hline
        \multicolumn{3}{c}{\thead{\textbf{Figure 6:} Context-free grammar for the main auxiliary dataset.}} 
    \end{tabular}
    }
    \label{cfg:main_aux}
\end{figure}

\begin{figure}[H]
    \centering
    \resizebox{0.95\textwidth}{!}{
    \begin{tabular}{lcl}
        \hline
        \thead[l]{S} & $\rightarrow$ & \thead[l]{NP\textsubscript{M} VP} \\
        \thead[l]{NP\textsubscript{M}} & $\rightarrow$ & \thead[l]{Det MNom $\mid$ Det MNom Prep Det Nom $\mid$ Det MNom RC} \\
        \thead[l]{NP\textsubscript{O}} & $\rightarrow$ & \thead[l]{Det Nom $\mid$ Det Nom Prep Det Nom $\mid$ Det Nom RC} \\
        \thead[l]{VP} & $\rightarrow$ & \thead[l]{Aux VI $\mid$ Aux VT NP\textsubscript{O}} \\
        \thead[l]{RC} & $\rightarrow$ & \thead[l]{Rel Aux VI $\mid$ Rel Det Nom Aux VT $\mid$ Rel Aux VT Det Nom} \\
        \thead[l]{Nom} & $\rightarrow$ & \thead[l]{N $\mid$ JJ Nom} \\
        \thead[l]{MNom} & $\rightarrow$ & \thead[l]{MNom1 $\mid$ MNom2} \\
        \thead[l]{MNom1} & $\rightarrow$ & \thead[l]{N $\mid$ JJ MNom1} \\
        \thead[l]{MNom2} & $\rightarrow$ & \thead[l]{N $\mid$ JJ MNom2 $\mid$ NS Poss MNom2 $\mid$ Nadj+MN} \\
        \thead[l]{Det} & $\rightarrow$ & \thead[l]{the $\mid$ some $\mid$ my $\mid$ your $\mid$ our $\mid$ her} \\
        \thead[l]{Poss} & $\rightarrow$ & \thead[l]{'s} \\
        \thead[l]{NS} & $\rightarrow$ & \thead[l]{bird $\mid$ bee $\mid$ ant $\mid$ duck $\mid$ lion $\mid$ dog $\mid$ tiger $\mid$ worm $\mid$ horse $\mid$ cat $\mid$ fish $\mid$ bear $\mid$ wolf} \\
        \thead[l]{N} & $\rightarrow$ & \thead[l]{bird $\mid$ bee $\mid$ ant $\mid$ duck $\mid$ lion $\mid$ dog $\mid$ tiger $\mid$ worm $\mid$ horse $\mid$ cat $\mid$ fish $\mid$ bear $\mid$ wolf $\mid$ birds $\mid$ bees $\mid$ ants $\mid$ \\ ducks $\mid$ lions $\mid$ dogs $\mid$ tigers $\mid$ worms $\mid$ horses $\mid$ cats $\mid$ fish $\mid$ bears $\mid$ wolves} \\
        \thead[l]{Nadj+MN} & $\rightarrow$ & \thead[l]{worker bee $\mid$ worker ant $\mid$ race horse $\mid$ queen bee $\mid$ german dog $\mid$ house cat} \\
        \thead[l]{VI} & $\rightarrow$ & \thead[l]{cry $\mid$ smile $\mid$ sleep $\mid$ swim $\mid$ wait $\mid$ move $\mid$ change $\mid$ read $\mid$ eat} \\
        \thead[l]{VT} & $\rightarrow$ & \thead[l]{dress $\mid$ kick $\mid$ hit $\mid$ hurt $\mid$ clean $\mid$ love $\mid$ accept $\mid$ remember $\mid$ comfort} \\ 
        \thead[l]{Aux} & $\rightarrow$ & \thead[l]{can $\mid$ will $\mid$ would $\mid$ could} \\
        \thead[l]{Prep} & $\rightarrow$ & \thead[l]{around $\mid$ near $\mid$ with $\mid$ upon $\mid$ by $\mid$ behind $\mid$ above $\mid$ below} \\
        \thead[l]{Rel} & $\rightarrow$ &\thead[l]{who $\mid$ that} \\
        \thead[l]{JJ} & $\rightarrow$ & \thead[l]{small $\mid$ little $\mid$ big $\mid$ hot $\mid$ cold $\mid$ good $\mid$ bad $\mid$ new $\mid$ old $\mid$ young} \\\hline
        \multicolumn{3}{c}{\thead{\textbf{Figure 7:} Context-free grammar for the subject noun dataset.}} 
    \end{tabular}
    }
    \label{cfg:sbjn}
\end{figure}

\vfill\clearpage

\begin{figure}[H]
    \centering
    \resizebox{0.95\textwidth}{!}{
    \begin{tabular}{lcl}
        \hline
        \thead[l]{S} & $\rightarrow$ & 
            \thead[l]{
            NP\textsubscript{sg\_Agr} Aux\textsubscript{sg} VI $\mid$ 
            NP\textsubscript{pl\_Agr} Aux\textsubscript{pl} VI 
            } \\
        \thead[l]{NP\textsubscript{sg\_Agr}} & $\rightarrow$ & 
            \thead[l]{
            Det N\textsubscript{sg} $\mid$ 
            Det N\textsubscript{sg} Prep Det N $\mid$ 
            Det N\textsubscript{sg} Prep RC\textsubscript{sg}
            } \\
        \thead[l]{NP\textsubscript{pl\_Agr}} & $\rightarrow$ & 
            \thead[l]{
            Det N\textsubscript{pl} $\mid$ 
            Det N\textsubscript{pl} Prep Det N $\mid$ 
            Det N\textsubscript{pl} Prep RC\textsubscript{pl}
            } \\
        \thead[l]{RC\textsubscript{sg}} & $\rightarrow$ & 
            \thead[l]{
            Rel Aux\textsubscript{sg} VI $\mid$
            Rel Aux\textsubscript{sg} VT Det N $\mid$
            Rel Det N\textsubscript{sg} Aux\textsubscript{sg} VT $\mid$
            Rel Det N\textsubscript{pl} Aux\textsubscript{pl} VT
            } \\
        \thead[l]{N} & $\rightarrow$ & \thead[l]{N\textsubscript{sg} $\mid$ N\textsubscript{pl}} \\
        \thead[l]{RC\textsubscript{pl}} & $\rightarrow$ & 
            \thead[l]{
            Rel Aux\textsubscript{pl} VI $\mid$
            Rel Aux\textsubscript{pl} VT Det N $\mid$
            Rel Det N\textsubscript{sg} Aux\textsubscript{sg} VT $\mid$
            Rel Det N\textsubscript{pl} Aux\textsubscript{pl} VT 
            } \\
        \thead[l]{Aux\textsubscript{sg}} & $\rightarrow$ & \thead[l]{does $\mid$ Modal} \\
        \thead[l]{Aux\textsubscript{pl}} & $\rightarrow$ & \thead[l]{do $\mid$ Modal} \\
        \thead[l]{Det} & $\rightarrow$ & \thead[l]{the $\mid$ some $\mid$ my $\mid$ your $\mid$ our $\mid$ her} \\
        \thead[l]{N\textsubscript{sg}} & $\rightarrow$ & \thead[l]{bird $\mid$ bee $\mid$ ant $\mid$ duck $\mid$ lion $\mid$ dog $\mid$ tiger $\mid$ worm $\mid$ horse $\mid$ cat $\mid$ fish $\mid$ bear $\mid$ wolf} \\
        \thead[l]{N\textsubscript{pl}} & $\rightarrow$ & \thead[l]{birds $\mid$ bees $\mid$ ants $\mid$ ducks $\mid$ lions $\mid$ dogs $\mid$ tigers $\mid$ worms $\mid$ horses $\mid$ cats $\mid$ fish $\mid$ bears $\mid$ wolves} \\
        \thead[l]{VI} & $\rightarrow$ & \thead[l]{cry $\mid$ smile $\mid$ sleep $\mid$ swim $\mid$ wait $\mid$ move $\mid$ change $\mid$ read $\mid$ eat} \\
        \thead[l]{VT} & $\rightarrow$ & \thead[l]{dress $\mid$ kick $\mid$ hit $\mid$ hurt $\mid$ clean $\mid$ love $\mid$ accept $\mid$ remember $\mid$ comfort} \\ 
        \thead[l]{VS} & $\rightarrow$ & \thead[l]{think $\mid$ say $\mid$ hope $\mid$ know} \\
        \thead[l]{VD} & $\rightarrow$ & \thead[l]{tell $\mid$ convince $\mid$ persuade $\mid$ inform} \\
        \thead[l]{Modal} & $\rightarrow$ & \thead[l]{can $\mid$ will $\mid$ would $\mid$ could} \\
        \thead[l]{Prep} & $\rightarrow$ & \thead[l]{around $\mid$ near $\mid$ with $\mid$ upon $\mid$ by $\mid$ behind $\mid$ above $\mid$ below} \\
        \thead[l]{Rel} & $\rightarrow$ &\thead[l]{who $\mid$ that} \\\hline
        \multicolumn{3}{c}{\thead{\textbf{Figure 8:} Context-free grammar for the subject-verb agreement dataset.}} 
    \end{tabular}
    }
    \label{cfg:agr}
\end{figure}

\begin{figure}[H]
    \centering
    \resizebox{0.95\textwidth}{!}{
    \begin{tabular}{lcl}
        \hline
        \thead[l]{S} & $\rightarrow$ & 
            \thead[l]{
            NP\textsubscript{M\_Ant} Aux VT Refl\textsubscript{M} $\mid$ 
            NP\textsubscript{F\_Ant} Aux VT Refl\textsubscript{F} $\mid$  \\  
            NP\textsubscript{M\_Ant} Aux VT Det N\textsubscript{F} by Refl\textsubscript{M} $\mid$ 
            NP\textsubscript{F\_Ant} Aux VT Det N\textsubscript{M} by Refl\textsubscript{F} $\mid$  \\
            NP\textsubscript{M\_Ant} Aux VT Det N\textsubscript{M} by Refl\textsubscript{M} $\mid$
            NP\textsubscript{F\_Ant} Aux VT Det N\textsubscript{F} by Refl\textsubscript{F}
            } \\
        \thead[l]{NP\textsubscript{M\_Ant}} & $\rightarrow$ & \thead[l]{Det N\textsubscript{M} $\mid$ Det N\textsubscript{M} RC} \\
        \thead[l]{NP\textsubscript{F\_Ant}} & $\rightarrow$ & \thead[l]{Det N\textsubscript{F} $\mid$ Det N\textsubscript{F} RC} \\
        \thead[l]{N} & $\rightarrow$ & \thead[l]{N\textsubscript{M} $\mid$ N\textsubscript{F}} \\
        \thead[l]{RC} & $\rightarrow$ & \thead[l]{Rel Aux VI $\mid$ Rel Det N Aux VT $\mid$ Rel Aux VT Det N} \\
        \thead[l]{Refl\textsubscript{M}} & $\rightarrow$ & \thead[l]{himself} \\
        \thead[l]{Refl\textsubscript{F}} & $\rightarrow$ & \thead[l]{herself} \\
        \thead[l]{Det} & $\rightarrow$ & \thead[l]{the $\mid$ some $\mid$ my $\mid$ your $\mid$ our $\mid$ her} \\
        \thead[l]{N\textsubscript{F}} & $\rightarrow$ & \thead[l]{girl $\mid$ woman $\mid$ queen $\mid$ actress $\mid$ sister $\mid$ wife $\mid$ mother $\mid$ princess $\mid$ aunt $\mid$ lady $\mid$ witch $\mid$ niece $\mid$ \\ nun} \\
        \thead[l]{N\textsubscript{M}} & $\rightarrow$ & \thead[l]{boy $\mid$ man $\mid$ king $\mid$ actor $\mid$ brother $\mid$ husband $\mid$ father $\mid$ prince $\mid$ uncle $\mid$ lord $\mid$ wizard $\mid$ nephew $\mid$ \\ monk} \\
        \thead[l]{VI} & $\rightarrow$ & \thead[l]{cry $\mid$ smile $\mid$ sleep $\mid$ swim $\mid$ wait $\mid$ move $\mid$ change $\mid$ read $\mid$ eat} \\
        \thead[l]{VT} & $\rightarrow$ & \thead[l]{dress $\mid$ kick $\mid$ hit $\mid$ hurt $\mid$ clean $\mid$ love $\mid$ accept $\mid$ remember $\mid$ comfort} \\ 
        \thead[l]{VS} & $\rightarrow$ & \thead[l]{think $\mid$ say $\mid$ hope $\mid$ know} \\
        \thead[l]{VD} & $\rightarrow$ & \thead[l]{tell $\mid$ convince $\mid$ persuade $\mid$ inform} \\
        \thead[l]{Aux} & $\rightarrow$ & \thead[l]{can $\mid$ will $\mid$ would $\mid$ could} \\
        \thead[l]{Prep} & $\rightarrow$ & \thead[l]{around $\mid$ near $\mid$ with $\mid$ upon $\mid$ by $\mid$ behind $\mid$ above $\mid$ below} \\
        \thead[l]{Rel} & $\rightarrow$ &\thead[l]{who $\mid$ that} \\\hline
        \multicolumn{3}{c}{\thead{\textbf{Figure 9:} Context-free grammar for the reflexive anaphora dataset.}} 
    \end{tabular}
    }
    \label{cfg:refl}
\end{figure}





\end{document}